\documentclass{article}

\usepackage{microtype}
\usepackage{graphicx}
\usepackage{subfigure}
\usepackage{booktabs} 
\usepackage{amsmath}
\usepackage{amssymb}
\usepackage{adjustbox}
\usepackage{float}
\usepackage{todonotes}
\usepackage{arydshln}
\usepackage{lipsum}
\usepackage{enumitem}

\allowdisplaybreaks

\usepackage[accepted]{icml2021}

\usepackage[colorlinks=true]{hyperref}

\icmltitlerunning{Opening the Blackbox: Accelerating Neural Differential Equations by Regularizing Internal Solver Heuristics}

\begin{document}

\twocolumn[
\icmltitle{Opening the Blackbox: Accelerating Neural Differential Equations by Regularizing Internal Solver Heuristics}




\begin{icmlauthorlist}
\icmlauthor{Avik Pal}{iit,jc}
\icmlauthor{Yingbo Ma}{jc}
\icmlauthor{Viral Shah}{jc}
\icmlauthor{Christopher Rackauckas}{jc,mit,pumas,umb}
\end{icmlauthorlist}

\icmlaffiliation{iit}{Indian Institute of Technology Kanpur}
\icmlaffiliation{mit}{Massachusetts Institute of Technology}
\icmlaffiliation{umb}{University of Maryland Baltimore}
\icmlaffiliation{pumas}{Pumas AI}
\icmlaffiliation{jc}{Julia Computing}

\icmlcorrespondingauthor{Avik Pal}{avikpal@cse.iitk.ac.in}
\icmlcorrespondingauthor{Christopher Rackauckas}{crackauc@mit.edu}

\icmlkeywords{Machine Learning, Neural Differential Equations, Regularization, Deep Learning, Time Series}

\vskip 0.3in
]



\printAffiliationsAndNotice{}  
\vspace*{-1em}
\begin{abstract}
\vspace*{-0.5em}
Democratization of machine learning requires architectures that automatically adapt to new problems. Neural Differential Equations (NDEs) have emerged as a popular modeling framework by removing the need for ML practitioners to choose the number of layers in a recurrent model. While we can control the computational cost by choosing the number of layers in standard architectures, in NDEs the number of neural network evaluations for a forward pass can depend on the number of steps of the adaptive ODE solver. But, can we force the NDE to learn the version with the least steps while not increasing the training cost? Current strategies to overcome slow prediction require high order automatic differentiation, leading to significantly higher training time. We describe a novel regularization method that uses the internal cost heuristics of adaptive differential equation solvers combined with discrete adjoint sensitivities to guide the training process towards learning NDEs that are easier to solve. This approach opens up the blackbox numerical analysis behind the differential equation solver's algorithm and directly uses its local error estimates and stiffness heuristics as cheap and accurate cost estimates. We incorporate our method without any change in the underlying NDE framework and show that our method extends beyond Ordinary Differential Equations to accommodate Neural Stochastic Differential Equations. We demonstrate how our approach can halve the prediction time and, unlike other methods which can increase the training time by an order of magnitude, we demonstrate similar reduction in training times. Together this showcases how the knowledge embedded within state-of-the-art equation solvers can be used to enhance machine learning.
\end{abstract}


\vspace{-2em}
\section{Introduction}
\label{sec:intro}

\begin{figure}[t]
    \centering
    \includegraphics[width=0.9\linewidth]{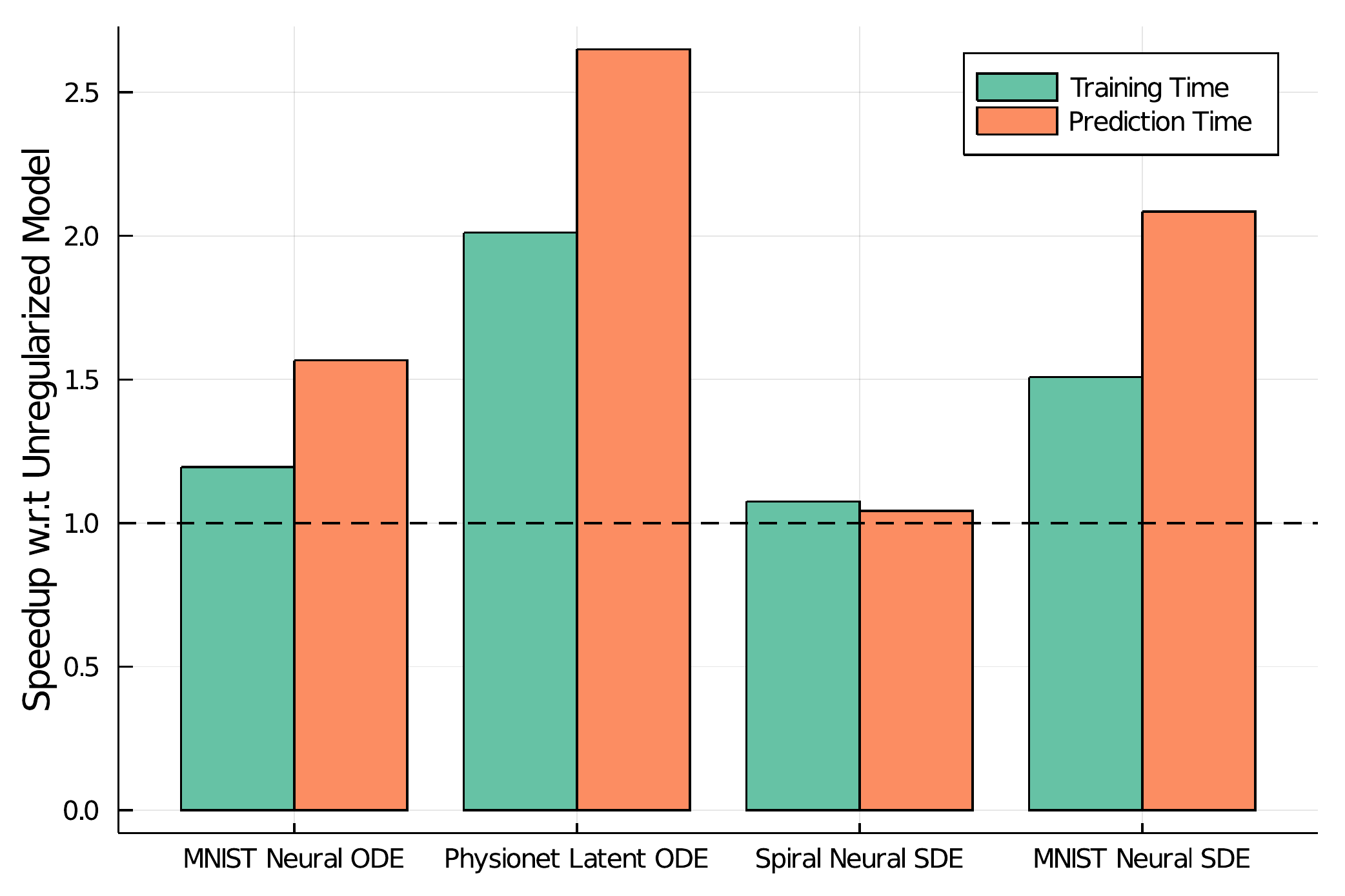}
    \vspace{-1.2em}
    \caption{\textbf{Training and Prediction Performance of Regularized NDEs} We obtain an average training and prediction speedup of $1.45$x and $1.84$x respectively for our best model on supervised classification and time series problems.}
    \vspace{-1em}
    \label{fig:performance}
\end{figure}

\begin{figure}[t]
    \centering
    \includegraphics[width=0.9\linewidth]{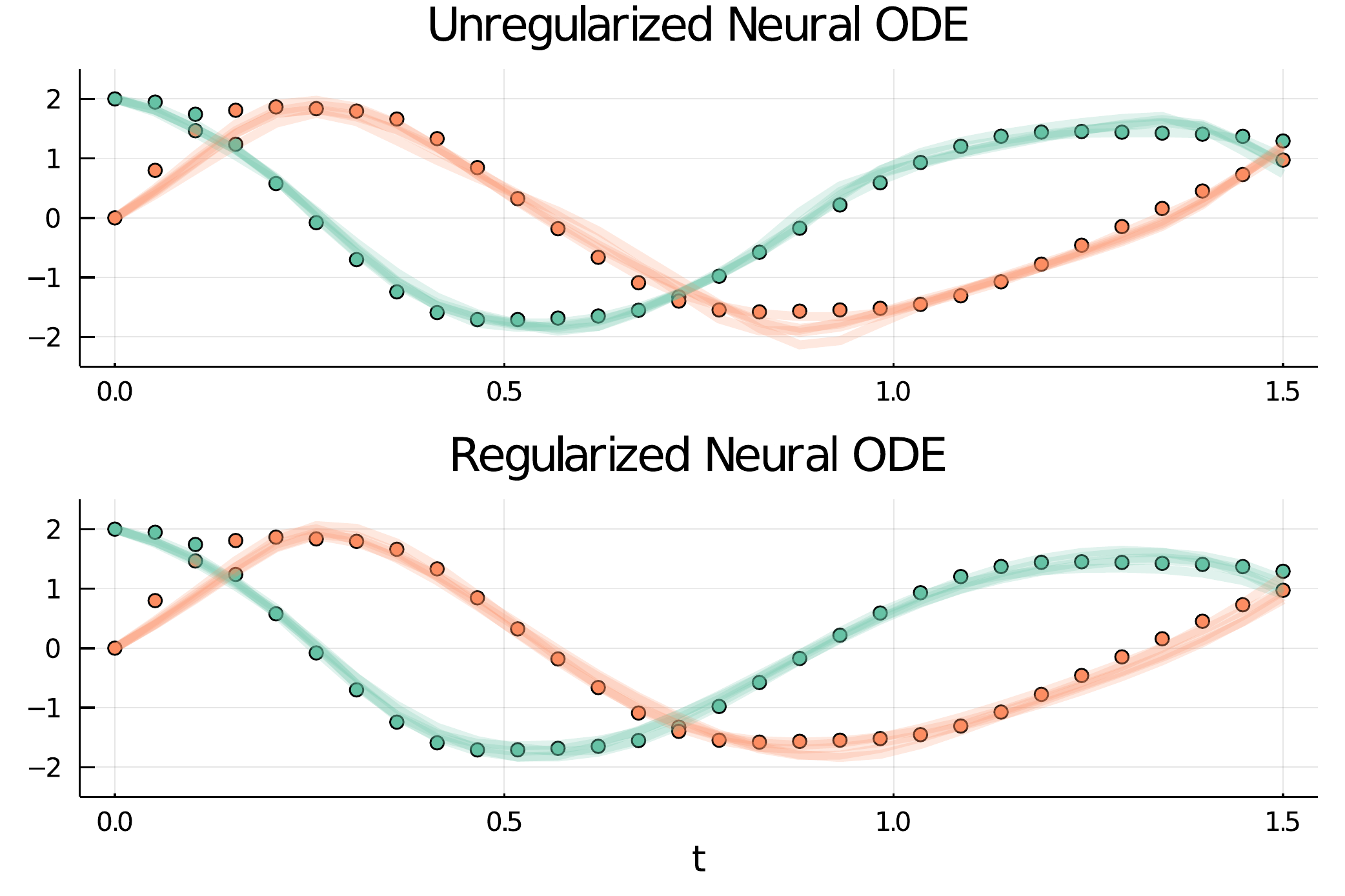}
    \vspace{-1.2em}
    \caption{\textbf{Error and Stiffness Regularization Keeps Accuracy.} We show the fits of the unregularized/regularized Neural ODE variants on the Sprial equation. However, the unregularized variant requires $1083.0 \pm 57.55$ NFEs while the one regularized using the stiffness and error estimates requires only $676.2 \pm 68.20$ NFEs, reducing prediction time by nearly 50\%.}
    \vspace{-1em}
    \label{fig:motivation}
\end{figure}

How many hidden layers should you choose in your recurrent neural network? \citet{chen2018neural} showed that the answer could be found automatically by using a continuous reformulation, the neural ordinary differential equation, and allowing an adaptive ODE solver to effectively choose the number of steps to take. Since then the idea was generalized to other domains such as stochastic differential equations \cite{liu2019neural, rackauckas2020universal} but one fact remained: solving a neural differential equation is expensive, and training a neural differential equation is even more so. In this manuscript we show a generally applicable method to force the neural differential equation training process to choose the least expensive option. We open the blackbox and show how using the numerical heuristics baked inside of these sophisticated differential equation solver codes allows for identifying the cheapest equations without requiring extra computation.


Our main contributions include:
\begin{itemize}
    \item We introduce a novel regularization scheme for neural differential equations based on the local error estimates and stiffness estimates. We observe that by white-boxing differential equation solvers to leverage pre-computed statistics about the neural differential equations, we can obtain faster training and prediction time while having a minimal effect on testing metrics.
    \item We compare our method with various regularization schemes~\citep{kelly2020learning, ghosh2020steer}, which often use higher order derivatives and are difficult to incorporate within existing systems. We empirically show that regularization using cheap statistics can lead to as efficient predictions as the ones requiring higher order automatic differentiation~\citep{kelly2020learning, finlay2020train} without the increased training time.
    \item We release our code\footnote{\url{https://github.com/avik-pal/RegNeuralODE.jl}}, implemented using the Julia Programming Language~\cite{Julia-2017} and SciML Software Suite~\cite{rackauckas2019diffeqflux}, with the intention of wider adoption of the proposed methods in the community.
\end{itemize}

\section{Background}
\label{sec:background}

\subsection{Neural Ordinary Differential Equations}

Ordinary Differential Equations (ODEs) are used to model the instantaneous rate of change ($\frac{dz(t)}{dt}$) of a state $z(t)$. Initial Value Problems (IVPs) are a class of ODEs that involve finding the state at a later time $t_1$, given the value $z_0$ at time $t_0$. This state, $z(t_1) = z_0 + \int_{t_0}^{t_1} f_\theta(z(t), t) dt$, generally cannot be computed analytically and requires numerical solvers. \citet{lu2018beyond} observed the similarity between fixed time-step discretization of ODEs and Residual Neural Networks~\citep{he2015deep}. \citet{chen2018neural} proposed the Neural ODE framework which use neural networks to model the ODE dynamics $\frac{dz(t)}{dt} = f_{\theta}(z(t), t)$. Using adaptive time stepping allows the model to operate at a variable continuous depth depending on the inputs. Removal of the fixed depth constraint of Residual Networks provides a more expressive framework and offer several advantages in problems like density estimation~\cite{grathwohl2018ffjord}, irregularly spaced time series problems~\cite{rubanova2019latent}, etc.

\subsection{Neural Stochastic Differential Equations}

Stochastic Differential Equations (SDEs) couple the effect of noise to a deterministic system of equations. SDEs are popularly used to model fluctuating stock prices, thermal fluctuations in physical systems, etc. In this paper, we only discuss SDEs with Diagonal Multiplicative Noise, though our method trivially extends to all other forms of SDEs. \citet{liu2019neural} propose an extension to Neural ODEs by stochastic noise injection in the form of Neural SDEs. Neural SDEs jointly train two neural networks $f_{\theta}$ and $g_{\phi}$, such that, the dynamics $dz(t) = f_{\theta}(z(t), t)dt + g_{\phi}(z(t), t)dW$. Stochastic Noise Injection regularize the training of continuous neural models and achieves significantly better robustness and generalization performance.

\subsection{Regularizing Neural ODEs for Speed}

Given the map $z(0)\rightarrow z(1)$ does not uniquely define the dynamics, it is possible to regularize the training process to learn differential equations that can be solved using fewer evaluations of $f_{\theta}$. In the case of continuous normalizing flows (CNF), the ordinary differential equation:
\begin{align}
    \frac{dz(t)}{dt} &= f_{\theta}(z(t), t) \\
    \frac{dy(t)}{dt} &= -\text{tr}\left(\frac{df_\theta}{dz}\right)
\end{align}
where $y(t)$ evolves a log-density~\cite{chen2018neural}. The FFJORD method improves the speed of CNF evaluations by approximating $\textit{tr}(\frac{df_\theta}{dz})$ via the Hutcheson trace estimator, i.e. $\textit{tr}(\frac{df_\theta}{dz}) = \mathbb{E}[\epsilon^T \frac{df_\theta}{dz} \epsilon]$ where $\epsilon \sim \mathcal{N}(0,1)$~\cite{hutchinson1989stochastic,grathwohl2018ffjord}. Subsequent research showed that this trace estimator could be used to regularize the Frobenius norm of the Jacobian $\Vert \frac{df_\theta}{dz}\Vert = \epsilon^T \frac{df_\theta}{dz}$~\cite{finlay2020train}. While $\epsilon^T \frac{df_\theta}{dz}$ is computationally expensive as it requires a reverse mode automatic differentiation evaluation in the model (leading to higher order differentiation), in the specific case of FFJORD this term is already required and thus this estimate is a computationally-free regularizer.

It was later shown that this form of regularization can be extended beyond FFJORD by using higher order automatic differentiation \cite{kelly2020learning}. This was done by regularizing a heuristic for the local error estimate, namely $\mathcal{R}_K(\theta) = \int_{t_0}^{t_f}\Vert \frac{d^K z(t)}{dt^K}\Vert^2_2 dt$. The authors showed Taylor-mode automatic differentiation improves the efficiency of calculating this estimator to a $\mathcal{O}(k^2)$ cost where $k$ is the order of derivative that is required, though this still implies that obtaining the 5 derivatives requires is a significant computational increase. In fact, the authors noted that ``when we train with adaptive solvers we do not improve overall training time'', and in fact giving a 1.7x slower training time. In this manuscript we show that this is all the way up to 10x on the PhysioNet challenge problem.

Here we show how to arrive at a similar regularization heuristic that is applicable to all neural ODE applications with suitable adaptive ODE solvers and requires no higher order automatic differentiation. We will show that this form of regularization is able to significantly improve training times and generalizes to other architectures like neural SDEs.

\subsection{Adaptive Time Stepping using Local Error Estimates}\label{sec:local_error}

Runge-Kutta Methods~\cite{runge1895numerische, kutta1901beitrag} are widely used for numerically approximating the solutions of ordinary differential equations. They are given by a tableau of coefficients $\{A,c,b\}$ where the stages $s$ are combined to produce an estimate for the update at $t+h$:
\begin{align}
\label{eq:rk}
\begin{split}
k_s &=  f\left(t+c_s h, z(t) + \sum_{i=1}^{s} a_{si} k_i\right)\\
z(t+h) &= z(t) + h \sum_{i=1}^{s} b_i k_i
\end{split}
\end{align}
For adaptivity, many Runge-Kutta methods include an alternative linear combiner $\tilde{b}_i$ such that $\tilde{z}(t+h) = z(t) + h \sum_{i=1}^{s} \tilde{b}_i k_i$ gives rise to an alternative solution, typically with one order less convergence~\cite{wanner1996solving,fehlberg1968classical,dormand1980family,Tsit5}. A classic result from Richardson extrapolation shows that $E = \Vert\tilde{z}(t+h) - z(t+h)\Vert$ is an estimate of the local truncation error~\citep{ascher1998computer, hairer1}. The goal of adaptive step size methods is to choose a maximal step size $h$ for which this error estimate is below user requested error tolerances. Given the absolute tolerance $atol$ and relative tolerance $rtol$, the solver satisfies the following constraint for determining the time stepping:
\begin{equation}
E \leq \text{atol} + \max(|z(t)|, |z(t+h)|) \cdot \text{rtol}
\end{equation}
The proportion of the error against the tolerance is thus: 
\begin{equation}
q = \left\Vert\frac{E}{\text{atol} + \max(|z_n|, |z_{n + 1}|) \cdot \text{rtol}}\right\Vert
\end{equation}
If $q < 1$ then the proposed time step $h$ is accepted, else it is rejected and reduced. In either case, a proportional error control scheme (P-control) proposes $h_{\text{new}} = \eta qh$, while a standard PI-controller of explicit adaptive Runge-Kutta methods can be shown to be equivalent to using:
\begin{equation}
h_{\text{new}} = \eta q_{n-1}^\alpha q_n^\beta h
\end{equation}
where $\eta$ is the safety factor, $q_{n-1}$ denotes the error proportion of the previous step, and $(\alpha,\beta)$ are the tunable PI gain hyperparameters~\cite{wanner1996solving}. Similar embedded methods error estimation schemes have also been derived for stochastic Runge-Kutta integrators of SDEs \cite{rackauckas2017adaptive,rackauckas2020sosri}. 

\subsection{Stiffness Estimation}

While there is no precise definition of stiffness, the definition used in practice is ``stiff equations are problems for which explicit methods don't work''~\cite{wanner1996solving,shampine1979user}. A simplified stiffness index is given by:
\begin{equation}
    S = \text{max}\|Re(\lambda_i)\|
\end{equation}
where $\lambda_i$ are the eigenvalues of the local Jacobian matrix. We note that various measures of stiffness have been introduced over the years, all being variations of conditioning of the pseudospectra~\cite{shampine2007stiff,higham1993stiffness}. The difficulty in defining a stiffness metric is that in each case, some stiff systems like the classic Robertson chemical kinetics or excited Van der Pol equation may violate the definition, meaning all such definitions are (useful) heuristics. In particular, it was shown that for explicit Runge-Kutta methods satisfying $c_x = c_y$ for some internal step, the term
\begin{equation}
    \|\lambda\| \approx \left\Vert\frac{ f(t+c_x h,\sum_{i=1}^{s} a_{xi}) - f(t+c_y h,\sum_{i=1}^{s} a_{yi})}{\sum_{i=1}^{s} a_{xi} - \sum_{i=1}^{s} a_{yi}}\right\Vert
\end{equation}
serves as an estimate to $S$~\cite{shampine1977stiffness}. Since each of these terms are already required in the Runge-Kutta updates of Equation \ref{eq:rk}, this gives a computationally-free estimate. This estimate is thus found throughout widely used explicit Runge-Kutta implementations, such as by the dopri method (found in suites like SciPy and Octave) to automatically exit when stiffness is detected~\cite{wanner1996solving}, and by switching methods which automatically change explicit Runge-Kutta methods to methods more suitable for stiff equations~\citep{rackauckas2019confederated}.

\section{Method}
\label{sec:main_methods}

\subsection{Regularizing Local Error and Stiffness Estimates}
\label{subsec:eest_reg}

Section \ref{sec:local_error} describes how larger local error estimates $E$ lead to reduced step sizes and thus a higher overall cost in the neural ODE training and predictions. Given this, we propose regularizing the neural ODE training process by the total local error in order to learn neural ODEs with as large step sizes as possible. Thus we define the regularizing term:
\begin{equation}\label{eq:reg_E}
    R_{E} = \sum_j E_j |h_j|
\end{equation}
summing over $j$ the time steps of the solution. This was done by accumulating the $E_j$ from the internals of the time stepping process at the end of each step. We note that this is similar to the regularization proposed in \cite{kelly2020learning}, namely:
\begin{equation}\label{eq:reg_K}
    R_{K} = \int_{t_0}^{t_1} \left\|\frac{d^K z(t)}{dt^K}\right\| dt
\end{equation}
where integrating over the $K^{th}$ derivatives is proportional to the principle (largest) truncation error term of the Runge-Kutta method \cite{hairer1}. However, this formulation requires high order automatic differentiation (which then is layered with reverse-mode automatic differentiation) which can be an expensive computation \cite{zhang2008computing} while Equation \ref{eq:reg_E} requires no differentiation. 

Similarly, the stiffness estimates at each step can be summed as:
%
\begin{equation}\label{eq:reg_S}
    R_{S} = \sum_j S_j 
\end{equation}
giving a computational heuristic for the total stiffness of the equation. Notably, both of these estimates $E_j$ and $S_j$ are already computed during the course of a standard explicit Runge-Kutta solution, making the forward pass calculation of the regularization term computationally free.

\subsection{Adjoints of Internal Solver Estimates}
\label{sec:adjoints}

Notice that $E_j = \sum_{i=1}^s (b_i-\tilde{b_i} )k_i$ cannot be constructed directly from the $z(t_j)$ trajectory of the ODE's solution. More precisely, the $k_i$ terms are not defined by the continuous ODE but instead by the chosen steps of the solver method. Continuous adjoint methods for neural ODEs \cite{chen2018neural, zhuang2021mali} only define derivatives in terms of the ODE quantities. This is required in order exploit properties such as allowing different steps in reverse and reversibility for reduced memory, and in constructing solvers requiring fewer NFEs~\cite{kidger2020hey}. Indeed, computing the adjoint of each stage variable $k_i$ can be done, but is known as discrete sensitivity analysis and is known to be equivalent to automatic differentiation of the solver \cite{zhang2014fatode}. Thus to calculate the derivative of the solution simultaneously to the derivatives of the solver states, we used direct automatic differentiation of the differential equation solvers for performing the experiments \cite{innes2018don}. We note that discrete adjoints are known to be more stable than continuous adjoints \cite{zhang2014fatode} and in the context of neural ODEs have been shown to stabilize the training process leading to better fits \cite{gholami2019anode,onken2020discretize}. While more memory intensive than some forms of the continuous adjoint, we note that checkpointing methods can be used to reduce the peak memory \cite{dauvergne2006data}. We note that this is equivalent to backpropagation of a fixed time step discretization if the step sizes are chosen in advance, and verify in the example code that no additional overhead is introduced.

\section{Experiments}

In this section, we consider the effectiveness of regularizing Neural Differential Equations (NDEs) on their training and prediction timings. We consider the following baselines while evaluating our models:
%
\begin{enumerate}[itemsep=0.25em]
    \item \textbf{Vanilla Neural (O/S)DE} with discrete sensitivities.
    \item \textbf{STEER}: Temporal Regularization for Neural ODE models by stochastic sampling of the end time during training~\citep{ghosh2020steer}.
    \item \textbf{TayNODE}: Regularizing the $K^{th}$ order derivatives of the Neural ODEs~\cite{kelly2020learning}\footnote{We use the original code formulation of the TayNODE in order to ensure usage of the specially-optimized Taylor-mode automatic differentiation technique \cite{bettencourt2019taylor} in the training process. Given the large size of the neural networks, most of the compute time lies in optimized BLAS kernels which are the same in both implementations, meaning we do not suspect library to be a major factor in timing differences beyond the AD specifics.}.
\end{enumerate}
We test our regularization on four tasks -- supervised image classification (Section~\ref{subsec:classificationode}) and time series interpolation (Section~\ref{subsec:ts_interp}) using Neural ODE, and fitting Neural SDE (Section~\ref{subsec:fitneuralsde}) and supervised image classification using Neural SDE (Section~\ref{subsec:classificationsde}). We use DiffEqFlux~\cite{rackauckas2019diffeqflux} and Flux~\cite{innes2018fashionable} 
for our experiments.

\subsection{Neural Ordinary Differential Equations}

In the following experiments, we use a Runge Kutta 5(4) solver
~\cite{Tsit5} 
with absolute and relative tolerances of $1.4 \times 10^{-8}$ to solve the ODEs. To measure the prediction time, we use a test batch size equal to the training batch size.

\subsubsection{Supervised Classification}
\label{subsec:classificationode}

\begin{figure}[t]
    \centering
    \includegraphics[width=0.9\linewidth]{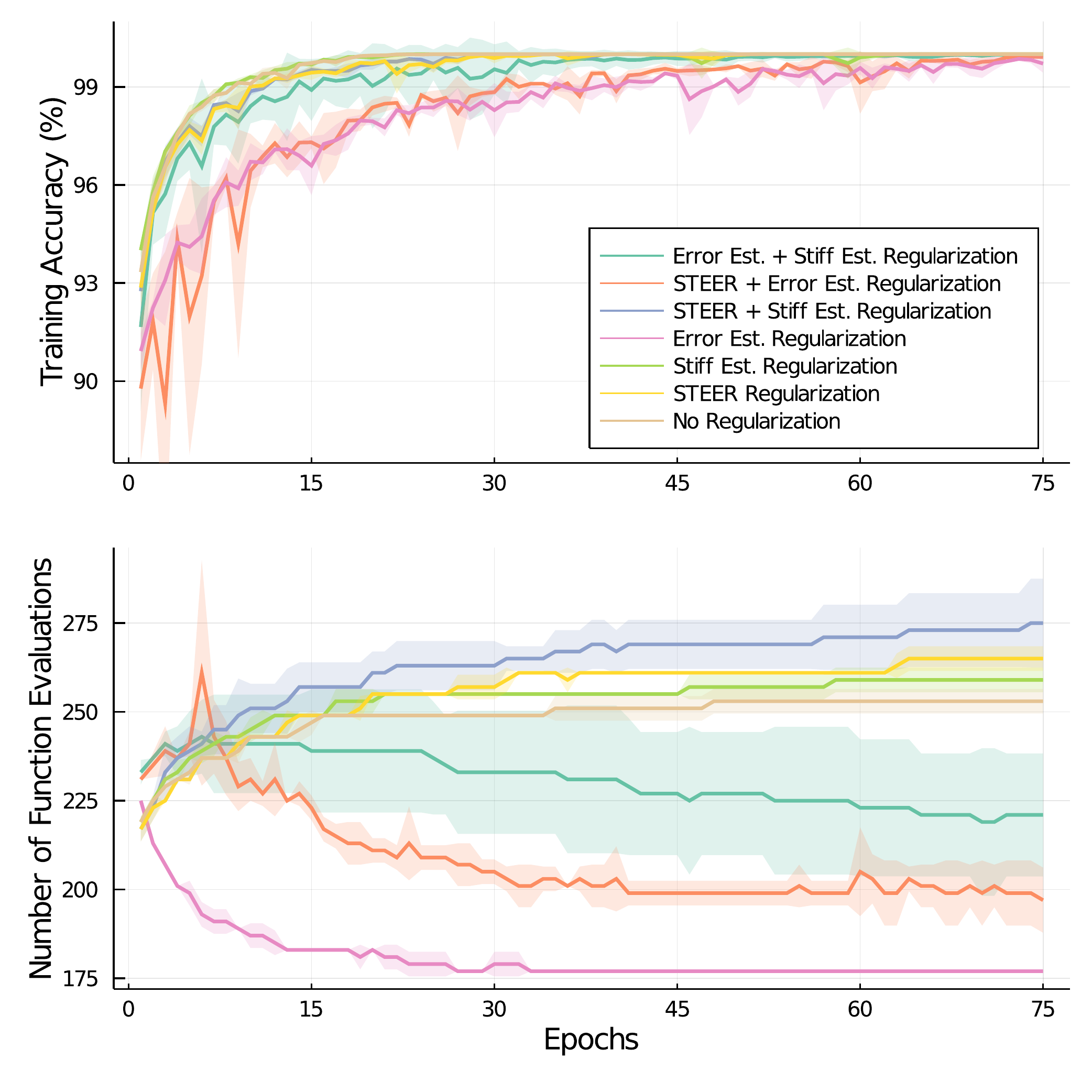}
    \vspace{-1.2em}
    \caption{\textbf{Number of Function Evaluations and Training Accuracy for Supervised MNIST Classification} Regularizing using ERNODE is the most consistent way to reduce the overall number of function evaluations. Using SRNODE alongside ERNODE stabilizes the training at the cost of increased prediction time.}
    \vspace{-1em}
    \label{fig:mnist_node}
\end{figure}

\begin{table*}[t]
    \centering
    \begin{adjustbox}{width=0.9\linewidth,center}
        \begin{tabular}{llllll}
            \toprule
            \textbf{Method} & \textbf{Train Accuracy (\%)} & \textbf{Test Accuracy (\%)} & \textbf{Train Time (hr)} & \textbf{Prediction Time (s)} & \textbf{NFE}\\
            \midrule
            Vanilla NODE & 100.0 $\pm$ 0.00 & 97.94 $\pm$ 0.02 & 0.98 $\pm$ 0.03 & 0.094 $\pm$ 0.010 & 253.0 $\pm$ 3.46\\
            STEER & 100.0 $\pm$ 0.00 & 97.94 $\pm$ 0.03 & 1.31 $\pm$ 0.07 & 0.092 $\pm$ 0.002 & 265.0 $\pm$ 3.46\\
            TayNODE & 98.98 $\pm$ 0.06 & 97.89 $\pm$ 0.00 & 1.19 $\pm$ 0.07 & 0.079 $\pm$ 0.007 & 080.3 $\pm$ 0.43\\
            \hdashline
            \textit{SRNODE (Ours)} & 100.0 $\pm$ 0.00 & 98.08 $\pm$ 0.15 & 1.24 $\pm$ 0.06 & 0.094 $\pm$ 0.003 & 259.0 $\pm$ 3.46\\
            \textit{ERNODE (Ours)} & 99.71 $\pm$ 0.28 & 97.32 $\pm$ 0.06 & 0.82 $\pm$ 0.02 & 0.060 $\pm$ 0.001 & 177.0 $\pm$ 0.00\\
            \hdashline
            STEER + \textit{SRNODE} & 100.0 $\pm$ 0.00 & 97.88 $\pm$ 0.06 & 1.55 $\pm$ 0.27 & 0.101 $\pm$ 0.009 & 275.0 $\pm$ 12.5\\
            STEER + \textit{ERNODE} & 99.91 $\pm$ 0.02 & 97.61 $\pm$ 0.11 & 1.37 $\pm$ 0.11 & 0.086 $\pm$ 0.018 & 197.0 $\pm$ 9.17\\
            \hdashline
            \textit{SRNODE} + \textit{ERNODE} & 99.98 $\pm$ 0.03 & 97.77 $\pm$ 0.05 & 1.37 $\pm$ 0.04 & 0.081 $\pm$ 0.006 & 221.0 $\pm$ 17.3\\
            \bottomrule
        \end{tabular}
    \end{adjustbox}
    \caption{\textbf{MNIST Image Classification using Neural ODE} Using ERNODE obtains a training and prediction speedup of 16.33\% and 37.78\% respectively, at only 0.6\% reduced prediction accuracy. SRNODE doesn't help in isolation but is effective when combined with ERNODE to reduce the prediction time by 14.44\% while incurring a reduced test accuracy of only 0.17\%.}
    \label{tab:mnist_node}
\end{table*}

\textbf{Training Details} We train a Neural ODE and a Linear Classifier to map flattened MNIST Images to their corresponding labels. Our model uses a two layered neural network $f_{\theta_1}$, as the ODE dynamics, followed by a linear classifier $g_{\theta_2}$, identical to the architecture used in \citet{kelly2020learning}.
\begin{align}
    z_{\theta_1}(x, t) &= \tanh(W_1 [x; t] + B_1)\\
    f_{\theta_1}(x, t) &= \tanh(W_2 [z_{\theta_1}(x, t); t] + B_2)\\
    g_{\theta_2}(x, t) &= \sigma(W_3 x + B_3)
\end{align}
where the parameters $W_1 \in \mathbb{R}^{100 \times 785}$, $B_1 \in \mathbb{R}^{100}$, $W_2 \in \mathbb{R}^{784 \times 101}$, $B_2 \in \mathbb{R}^{784}$, $W_3 \in \mathbb{R}^{10 \times 784}$, and $B_3 \in \mathbb{R}^{10}$. We use a batch size of $512$ and train the model for $75$ epochs using Momentum~\cite{qian1999momentum} with learning rate of $0.1$ and mass of $0.9$, and a learning rate inverse decay of $10^{-5}$ per iteration. For Error Estimate Regularization, we perform exponential annealing of the regularization coefficient from $100.0$ to $10.0$ over $75$ epochs. For Stiffness Regularization, we use a constant coefficient of $0.0285$.

\textbf{Baselines} For the STEER baseline, we train the models by stochastically sampling the end time point from $\mathcal{U}(T - b, T + b)$ where $T = 1.0$ and $b = 0.5$\footnote{$b=0.25$ was also considered but final results were comparable}. We observe no training improvement but there is a minor improvement in prediction time. For the TayNODE baseline, we train the model with a reduced batch size of 100\footnote{Batch Size was reduced to ensure we reach a comparable train/test accuracy as the other trained models.}, $\lambda = 3.02 \times 10^{-3}$, and regularizing $3^{rd}$ order derivatives.

\textbf{Results} Figure~\ref{fig:mnist_node} visualizes the training accuracy and number of function evaluations over training. Table~\ref{tab:mnist_node} summarizes the metrics from the trained baseline and proposed models -- Error Estimate Regularized Neural ODE (\textit{ERNODE}) and Stiffness Regularized Neural ODE (\textit{SRNODE}). Additionally, we perform ablation studies by composing various regularization strategies.


\subsubsection{Time Series Interpolation}
\label{subsec:ts_interp}

\begin{table*}[t]
    \centering
    \begin{adjustbox}{width=0.9\linewidth,center}
        \begin{tabular}{llllll}
            \toprule
            \textbf{Method} & \textbf{Train Loss ($\times 10^{-3}$)} & \textbf{Test Loss ($\times 10^{-3}$)} & \textbf{Train Time (hr)} & \textbf{Prediction Time (s)} & \textbf{NFE}\\
            \midrule
            Vanilla NODE & 3.48 $\pm$ 0.00 & 3.55 $\pm$ 0.00 & 1.75 $\pm$ 0.39 & 0.53 $\pm$ 0.12 & 733.0 $\pm$ 84.29 \\
            STEER & 3.43 $\pm$ 0.02 & 3.48 $\pm$ 0.01 & 1.62 $\pm$ 0.26 & 0.54 $\pm$ 0.06 & 699.0 $\pm$ 141.1\\
            TayNODE & 4.21 $\pm$ 0.02 & 4.21 $\pm$ 0.01 & 12.3 $\pm$ 0.32 & 0.22 $\pm$ 0.02 & 167.3 $\pm$ 11.93 \\
            \hdashline
            \textit{SRNODE (Ours)} & 3.52 $\pm$ 1.44 & 3.58 $\pm$ 0.05 & 0.87 $\pm$ 0.09 & 0.20 $\pm$ 0.01 & 273.0 $\pm$ 0.000\\
            \textit{ERNODE (Ours)} & 3.51 $\pm$ 0.00 & 3.57 $\pm$ 0.00 & 0.94 $\pm$ 0.13 & 0.21 $\pm$ 0.02 & 287.0 $\pm$ 17.32\\
            \hdashline
            STEER + \textit{SRNODE} & 3.67 $\pm$ 0.02 & 3.73 $\pm$ 0.02 & 0.89 $\pm$ 0.08 & 0.20 $\pm$ 0.01 & 271.0 $\pm$ 12.49\\
            STEER + \textit{ERNODE} & 3.41 $\pm$ 0.02 & 3.48 $\pm$ 0.01 & 1.03 $\pm$ 0.25 & 0.24 $\pm$ 0.05 & 269.0 $\pm$ 33.05 \\
            \hdashline
            \textit{SRNODE} + \textit{ERNODE} & 3.48 $\pm$ 0.11 & 3.56 $\pm$ 0.03 & 1.12 $\pm$ 0.08 & 0.21 $\pm$ 0.01 & 263.0 $\pm$ 12.49 \\
            \bottomrule
        \end{tabular}
    \end{adjustbox}
    \caption{\textbf{Physionet Time Series Interpolation} All the regularized variants of Latent ODE (except STEER) have comparable prediction times. Additionally, the training time is reduced by $36\% - 50\%$ on using one of our proposed regularizers, while TayNODE increases the training time by $7$x. Overall, SRNODE has the best training and prediction timings while incurring an increased $0.85\%$ test loss.}
    \label{tab:latent_ode}
\end{table*}

\textbf{Training Details} We use the Latent ODE~\citep{chen2018neural} model with RNN encoder to learn the trajectories for ICU Patients for Physionet Challenge 2012 Dataset~\citep{silva2012predicting}. We use the preprocessed data provided by \citet{kelly2020learning} to ensure consistency in results. For every independent run, we perform an $80:20$ split of the data for training and evaluation.

Our model architecture is similar to the encoder-decoder models used in \citet{rubanova2019latent}. We use a 20-dimensional latent state and a 40-dimensional hidden state for the recognition model. Our ODE dynamics is given by a 4-layered neural network with 50 units and tanh activation. We train our models for $300$ epochs with a batchsize of $512$ and using Adamax~\citep{kingma2017adam} with a learning rate of $0.01$ and an inverse decay of $10^{-5}$. We minimize the negative log likelihood of the predictions and perform KL annealing with a coefficient of $0.99$.

For Error Estimate Regularization, we perform exponential annealing of the regularization coefficient from $1000.0$ to $100.0$ over $300$ epochs. We note that using $R_{E} = \sum_j E_j^2$, instead of $R_{E} = \sum_j E_j |h_j|$, yields similar results with a constant regularization coefficient of $100.0$. For Stiffness Regularization, we use a constant coefficient of $0.285$.

\textbf{Baselines} For STEER Baseline, we stochastically sample the timestep to evaluate the difference between interpolated and ground truth data. Essentially for the interval $(t_i, t_{i + 1})$, we evaluate the model at $\mathcal{U}(t_{i + 1} - \frac{t_{i + 1} - t_i}{2}, t_{i + 1} + \frac{t_{i + 1} - t_i}{2})$ and compare with the truth at $t_{i + 1}$. We sample end points after every iteration of the model. STEER reduces the training time but has no significant effect on the prediction time. TayNODE was trained by regularizing the $2^{nd}$ order derivatives and a coefficient of $0.01$ for 300 epochs and a batchsize of $512$. TayNODE had an exceptionally high training time $\sim 7\times$ compared to the unregularized baseline.

\begin{figure}[t]
    \centering
    \includegraphics[width=0.9\linewidth]{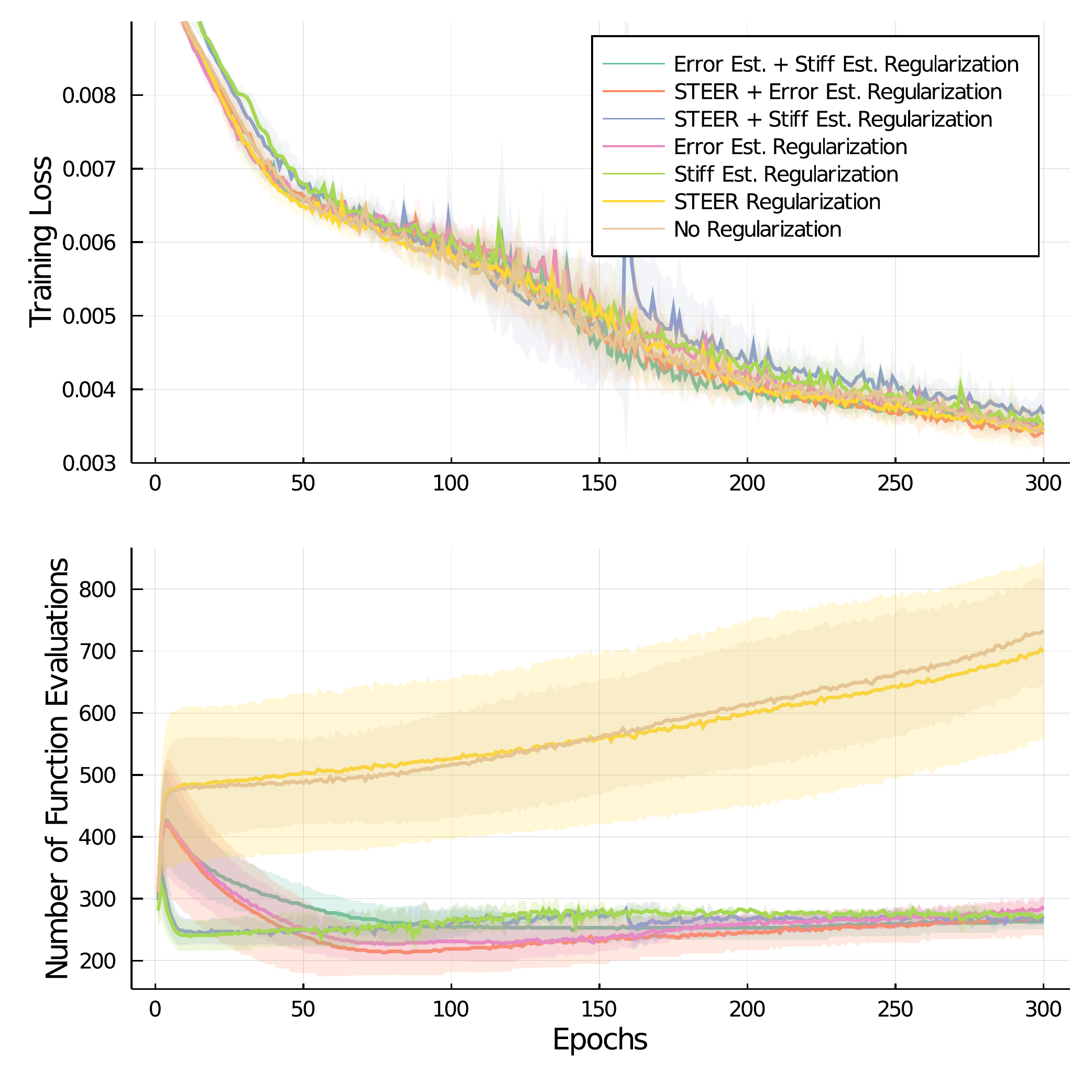}
    \vspace{-1.2em}
    \caption{\textbf{Number of Function Evaluations and Training Loss for Physionet Time Series Interpolation} Regularized and Unregularized variants of the model have very similar trajectories for the training loss. We do notice a significant difference in the NFE plot. Using either Error Estimate Regularization or Stiffness Regularization is able to bound the NFE to $< 300$, compared to $\sim 700$ for STEER or unregularized Latent ODE.}
    \vspace{-1em}
    \label{fig:latent_ode}
\end{figure}

\textbf{Results} Figure~\ref{fig:latent_ode} shows the training MSE loss and the NFE counts for the considered models. Table~\ref{tab:latent_ode} summarizes the metrics and wall clock timings for the baselines, proposed regularizers and their compositions with previously proposed regularizers. We observe that SRNODE provides the most significant speedup while ERNODE attains similar losses at slightly higher training and prediction times.

\subsection{Neural Stochastic Differential Equations}

In these experiments, we use SOSRI/SOSRI2~\citep{rackauckas2020sosri} to solve the Neural SDEs. The wall clock timings represent runs on a CPU.

\subsubsection{Fitting Spiral Differential Equation}
\label{subsec:fitneuralsde}

\begin{table*}[t]
    \centering
    \begin{adjustbox}{width=0.75\linewidth,center}
        \begin{tabular}{lllll}
            \toprule
            \textbf{Method} & \textbf{Mean Squared Loss} & \textbf{Train Time (s)} & \textbf{Prediction Time (s)} & \textbf{NFE}\\
            \midrule
            Vanilla NSDE & 0.0217 $\pm$ 0.0088 & 178.95 $\pm$ 20.22 & 0.07553 $\pm$ 0.0186 & 528.67 $\pm$ 6.11\\
            \hdashline
            \textit{SRNSDE (Ours)} & 0.0204 $\pm$ 0.0091 & 166.42 $\pm$ 14.51 & 0.07250 $\pm$ 0.0017 & 502.00 $\pm$ 4.00 \\
            \textit{ERNSDE (Ours)} & 0.0227 $\pm$ 0.0090 & 173.43 $\pm$ 04.18 & 0.07552 $\pm$ 0.0008 & 502.00 $\pm$ 4.00\\
            \bottomrule
        \end{tabular}
    \end{adjustbox}
    \caption{\textbf{Spiral SDE} The ERNSDE attains a relative loss of 4\% compared to vanilla Neural SDE while reducing the training time and number of function evaluations. Using SRNSDE reduces both the training and prediction times by 7\% and 4\% respectively.}
    \label{tab:fit_spiral_sde}
\end{table*}

\begin{figure}[t]
    \centering
    \includegraphics[width=0.9\linewidth]{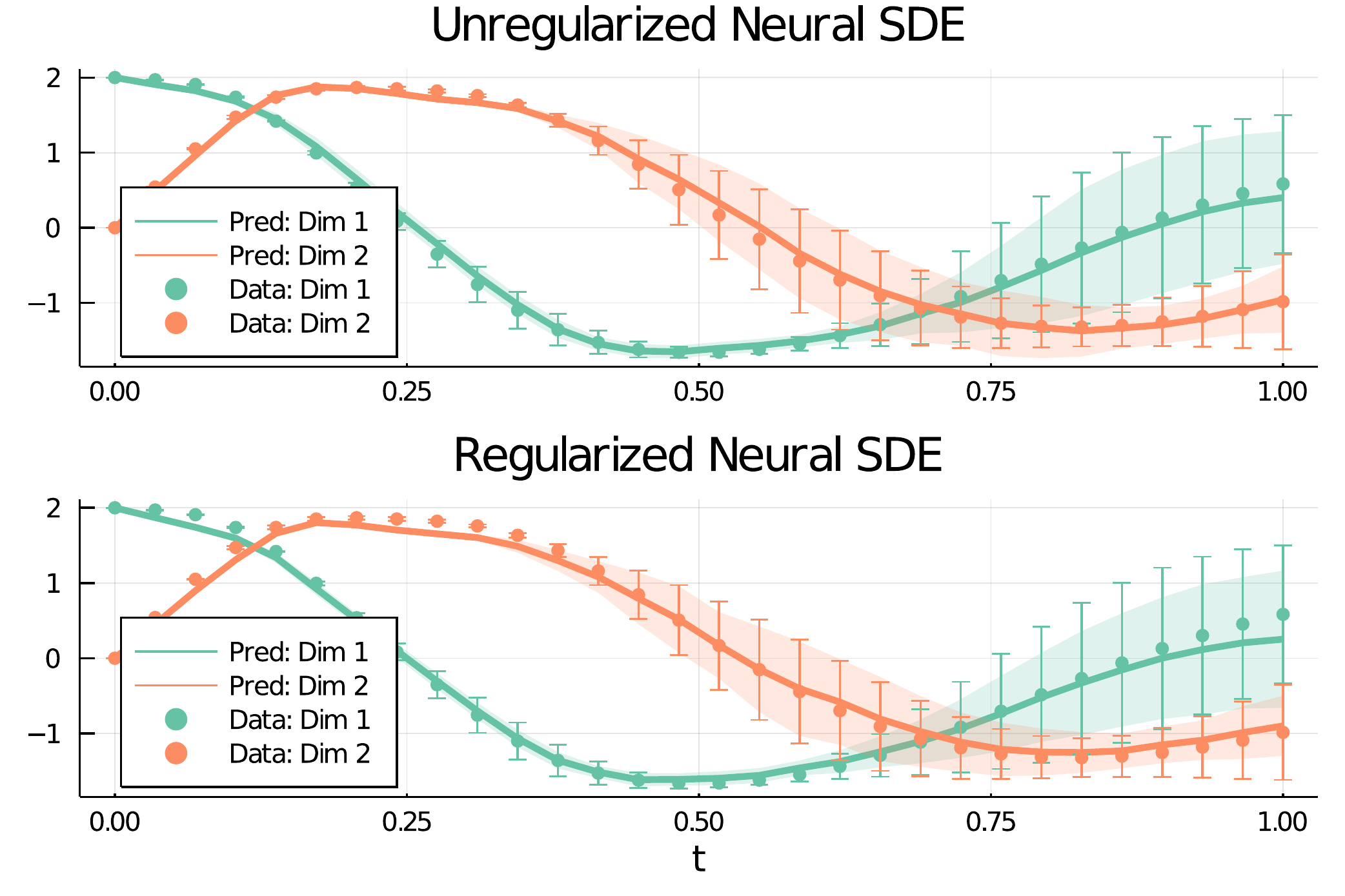}
    \vspace{-1.2em}
    \caption{\textbf{Fitting a Neural SDE on Spiral SDE Data.} Regularizing has minimal effect on the learned dynamics with reduced training and prediction cost.}
    \vspace{-1.2em}
    \label{fig:fit_neural_sde}
\end{figure}

\textbf{Training Details} In this experiment, we consider training a Neural SDE to mimic the dynamics of the Spiral Stochastic Differential Equation with Diagonal Noise (DSDE). Spiral DSDE is prescribed by the following equations:
\begin{align}
\begin{split}
    du_1 &= -\alpha u_1^3 dt + \beta u_2^3 dt + \gamma u_1 dW\\
    du_2 &= -\beta u_1^3 dt - \alpha u_2^3 dt + \gamma u_2 dW
\end{split}
\end{align}
where $\alpha = 0.1$, $\beta = 2.0$, and $\gamma = 0.2$. We
generate data across $10000$ trajectories at 30 uniformly spaced points between $t \in [0, 1]$ (Figure~\ref{fig:fit_neural_sde}). We parameterize our drift and diffusion functions using neural networks $f_\theta$ and $g_\phi$ via:
\begin{align}
\begin{split}
    f_\theta(x, t) &= W_2 \tanh(W_1 x^3 + B_1) + B_2\\
    g_\phi(x, t) &= W_3 x + B_3
\end{split}
\end{align}
where the parameters $W_1 \in \mathbb{R}^{50 \times 2}$, $B_1 \in \mathbb{R}^{50}$, $W_2 \in \mathbb{R}^{2 \times 50}$, $B_2 \in \mathbb{R}^{2}$, $W_3 \in \mathbb{R}^{2 \times 2}$, and $B_3 \in \mathbb{R}^{2}$. For fitting the drift and diffusion functions to the simulated data, we used a generalized method of moments loss function \cite{luck2016generalized,jeisman2006estimation}. Our objective is to train these parameters to minimize the $L_2$ distance between the mean ($\mu$) and variance ($\sigma^2$) of predicted and real data. Let, $\hat{\mu}_i$'s and $\hat{\sigma}^2_i$'s denote the means and variances respectively of the multiple predicted trajectories.
\begin{equation}
    \mathcal{L}(u_0; \theta, \phi) = \sum_{i = 1}^{30} \left[(\mu_i - \hat{\mu}_i)^2 + (\sigma^2_i - \hat{\sigma}^2_i)^2\right] + \lambda_r R_E
\end{equation}

The models were trained using AdaBelief Optimizer~\cite{zhuang2020adabelief} with a learning rate of $0.01$ for $250$ iterations. We generate 100 trajectories for each iteration to compute the $\hat{\mu}_i$s and $\hat{\sigma}^2_i$s.

\textbf{Results} Table~\ref{tab:fit_spiral_sde} summarizes the final results for the trained models for 3 different random seeds. We notice that even for this ``toy" problem, we can marginally improve training time while incurring a minimal penalty on the final loss.

\subsubsection{Supervised Classification}
\label{subsec:classificationsde}

\textbf{Training Details} We train a Neural SDE model to map flattened MNIST Images to their corresponding labels. Our diffusion function uses a two layered neural network $f_{\theta_2}$ and the drift function is a linear map $g_{\theta_3}$. We use two additional linear maps -- $a_{\theta_1}$ mapping the flattened image to the hidden dimension and $b_{\theta_4}$ mapping the output of the Neural SDE to the logits.
\begin{align}
    a_{\theta_1}(x, t) &= W_1 x + B_1\\
    f_{\theta_2}(x, t) &= W_3 ~ \tanh(W_2 ~ x + B_2) + B_3\\
    g_{\theta_3}(x, t) &= W_4 ~ x + B_4\\
    b_{\theta_4}(x, t) &= W_5 ~ x + B_5   
\end{align}
where the parameters $W_1 \in \mathbb{R}^{32 \times 784}$, $B_1 \in \mathbb{R}^{32}$, $W_2 \in \mathbb{R}^{32 \times 64}$, $B_2 \in \mathbb{R}^{64}$, $W_3 \in \mathbb{R}^{32 \times 64}$, $B_3 \in \mathbb{R}^{32}$, $W_4 \in \mathbb{R}^{10 \times 32}$, and $B_3 \in \mathbb{R}^{10}$. We use a batch size of $512$ and train the model for $40$ epochs using Adam~\cite{kingma2017adam} with learning rate of $0.01$, and an inverse decay of $10^{-5}$ per iteration. While making predictions we use the mean logits across $10$ trajectories. For Error Estimate and Stiffness Regularization, we use constant coefficients $10.0$ and $0.1$ respectively.

\begin{figure}[t]
    \centering
    \includegraphics[width=0.9\linewidth]{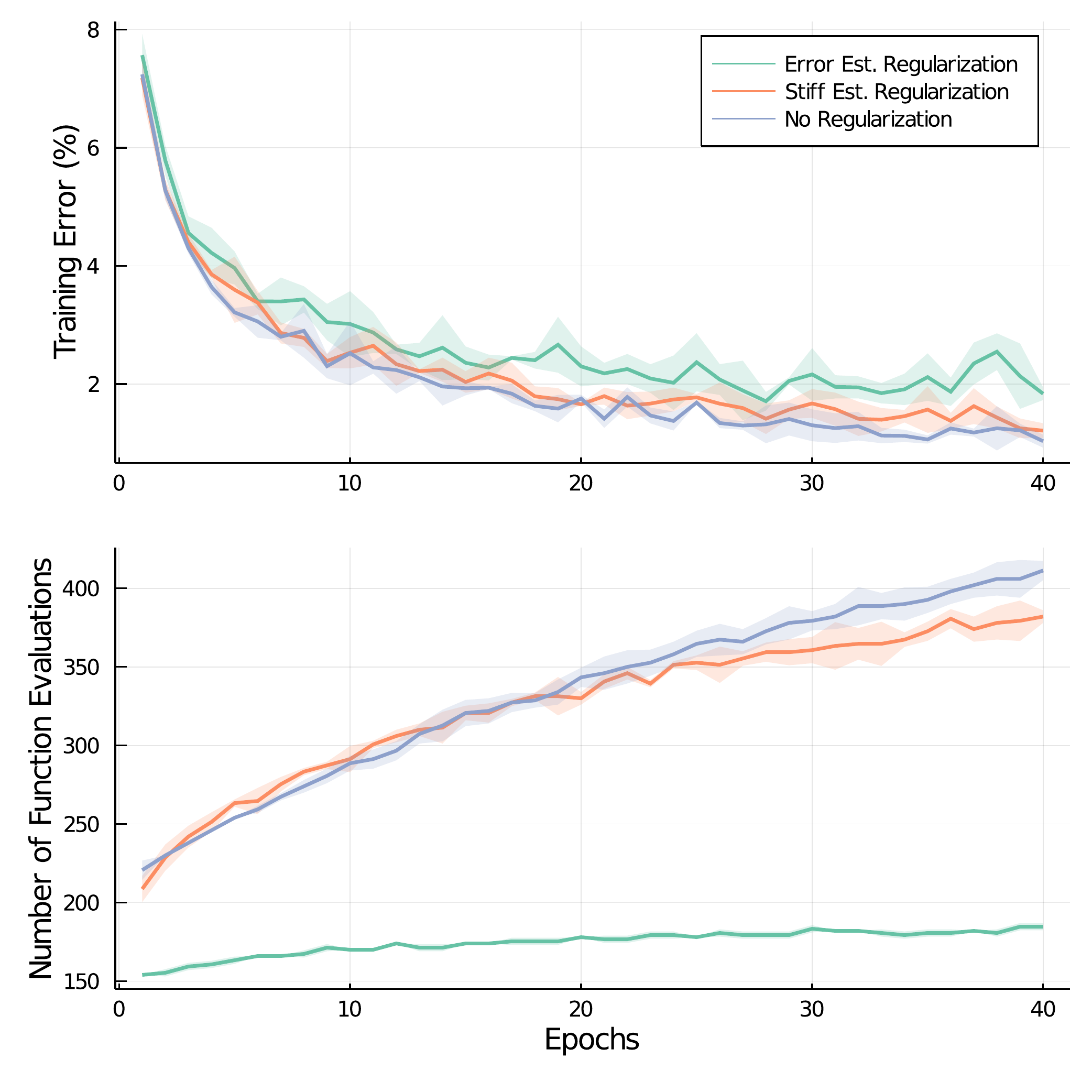}
    \vspace{-1.2em}
    \caption{\textbf{Number of Function Evaluations and Training Error for Supervised MNIST Classification using Neural SDE} ERNSDE reduces the NFE below 300 with minimal error change while the unregularized version has NFE $\sim 400$.}
    \vspace{-1em}
    \label{fig:mnist_nsde}
\end{figure}

\textbf{Results} Figure~\ref{fig:mnist_nsde} shows the variation in NFE and Training Error during training. Table~\ref{tab:mnist_nsde} summarizes the final metrics and timings for all the trained models. We observe that SRNSDE doesn't improve the training/prediction time, similar to the MNIST Neural ODE Experiment~\ref{subsec:classificationode}. However, ERNSDE gives us a training and prediction speedup of $33.7\%$ and $52.02\%$ respectively, at the cost of $0.7\%$ reduced test accuracy.

\begin{table*}[t]
    \centering
    \begin{adjustbox}{width=0.85\linewidth,center}
        \begin{tabular}{llllll}
            \toprule
            \textbf{Method} & \textbf{Train Accuracy (\%)} & \textbf{Test Accuracy (\%)} & \textbf{Train Time (hr)} & \textbf{Prediction Time (s)} & \textbf{NFE}\\
            \midrule
            Vanilla NSDE & 98.97 $\pm$ 0.11 & 96.95 $\pm$ 0.11 & 6.32 $\pm$ 0.19 & 15.07 $\pm$ 0.93 & 411.33 $\pm$ 6.11\\
            \hdashline
            \textit{SRNSDE (Ours)} & 98.79 $\pm$ 0.12 & 96.80 $\pm$ 0.07 & 8.54 $\pm$ 0.37 & 14.50 $\pm$ 0.40 & 382.00 $\pm$ 4.00\\
            \textit{ERNSDE (Ours)} & 98.16 $\pm$ 0.11 & 96.27 $\pm$ 0.35 & 4.19 $\pm$ 0.04 & 07.23 $\pm$ 0.14 & 184.67 $\pm$ 2.31\\
            \bottomrule
        \end{tabular}
    \end{adjustbox}
    \caption{\textbf{MNIST Image Classification using Neural SDE} ERNSDE obtains a training and prediction speedup of 33.7\% and 52.02\% respectively, at only 0.7\% reduced prediction accuracy.}
    \vspace{-1.5em}
    \label{tab:mnist_nsde}
\end{table*}

\vspace{-1em}
\section{Discussion}

Numerical analysis has had over a century of theoretical developments leading to efficient adaptive methods for solving many common nonlinear equations such as differential equations. Here we demonstrate that by using the knowledge embedded within the heuristics of these methods we can accelerate the training process of neural ODEs. 

We note that on the larger sized PhysioNet and MNIST examples we saw significant speedups while on the smaller differential equation examples we saw only minor performance improvements. This showcases how the NFE becomes a better estimate of the total compute time as the cost of the ODE $f$ (and SDE $g$) increase when the model size increases.

This result motivates efforts in differentiable programming \cite{wang2018backpropagation,abadi2019simple,rackauckas2020generalized} which enables direct differentiation of solvers since utilizing the solver's heuristics may be crucial in the development of advanced techniques. This idea could be straightforwardly extended not only to other forms of differential equations, but also to other ``implicit layer'' machine learning methods. For example, Deep Equilibrium Models (DEQ) \cite{bai2019deep} model the system as the solution to an implicit function via a nonlinear solver like Bryoden or Newton's method. Heuristics like the ratio of the residuals have commonly been used as a convergence criterion and as a work estimate for the difficulty of solving a particular nonlinear equation \cite{wanner1996solving}, and thus could similarly be used to regularize for learning DEQs whose forward passes are faster to solve. Similarly, optimization techniques such as BFGS \cite{kelley1999iterative} contain internal estimates of the Hessian which can be used to regularize the stiffness of ``optimization as layers'' machine learning architectures like OptNet \cite{amos2017optnet}. However, in these cases we note that continuous adjoint techniques have a significant computational advantage over discrete adjoint methods because the continuous adjoint method can be computed directly at the point of the solution while discrete adjoints would require differentiating through the iteration process. Thus while a similar regularization would exist in these contexts, in the case of differential equations the continuous and discrete adjoints share the same computational complexity which is not the case in methods which iterate to convergence. Further study of these applications would be required in order to ascertain the effectiveness in accelerating the training process, though by extrapolation one may guess that at least the forward pass would be accelerated.

\vspace{-1em}
\section{Limitations}

While these experiments have demonstrated major performance improvements, it is pertinent to point out the limitations of the method. One major point to note is that this only applies to learning neural ODEs for maps $z(0) \mapsto z(1)$ as is used in machine learning applications of the architecture \cite{chen2018neural}. Indeed, a neural ODE as an ``implicit layer'' for predictions in machine learning does not require identification of dynamical mechanisms. However, if the purpose is to learn the true governing dynamics a physical system from timeseries data, this form of regularization would bias the result, dampening higher frequency responses leading to an incorrect system identification. Approaches which embed neural networks into solvers could be used in such cases \cite{shen2020deep,poli2020hypersolvers}. Indeed we note that such Hypereuler approaches could be combined with the ERNODE regularization on machine learning prediction problems, which could be a fruitful avenue of research. Lastly, we note that while either the local error and stiffness regularization was effective on each chosen equation, neither was effective on all equations and at this time there does not seem to be a clear a priori indicator as to which regularization is necessary for a given problem. While it seems the error regularization was more effective on the image classification tasks while the stiffness regularization was more effective on the time series task, we believe more experiments will be required in order to ascertain whether this is a common phenomena, possibly worthy of theoretical investigation.

\vspace{-1em}
\section{Conclusion}

Our studies reveal that error estimate regularization provides a consistent way to improve the training/prediction time of neural differential equations. In our experiments, we see an average improvement of $1.4$x training time and $1.8$x prediction time on using error estimate regularization. Overall we provide conclusive evidence that cheap and accurate cost estimates obtained by white-boxing differential equation solvers can be as effective as expensive higher-order regularization strategies. Together these results demonstrate a generalizable idea for how to combine differentiable programming with algorithm heuristics to improve training speeds in a way that cannot be done with continuous adjoint techniques. Thus, even if a derivative can be defined for a given piece of code, our approach shows that differentiating the solver can still have major advantages because the solver internal details in terms of stability and performance.

\vspace{-1em}
\section{Acknowledgements}
This material is based upon work supported by the National Science Foundation under grant no. OAC-1835443, grant no. SII-2029670, grant no. ECCS-2029670, grant no. OAC-2103804, and grant no. PHY-2021825. We also gratefully acknowledge the U.S. Agency for International Development through Penn State for grant no. S002283-USAID. The information, data, or work presented herein was funded in part by the Advanced Research Projects Agency-Energy (ARPA-E), U.S. Department of Energy, under Award Number DE-AR0001211 and DE-AR0001222. We also gratefully acknowledge the U.S. Agency for International Development through Penn State for grant no. S002283-USAID. The views and opinions of authors expressed herein do not necessarily state or reflect those of the United States Government or any agency thereof. This material was supported by The Research Council of Norway and Equinor ASA through Research Council project "308817 - Digital wells for optimal production and drainage". Research was sponsored by the United States Air Force Research Laboratory and the United States Air Force Artificial Intelligence Accelerator and was accomplished under Cooperative Agreement Number FA8750-19-2-1000. The views and conclusions contained in this document are those of the authors and should not be interpreted as representing the official policies, either expressed or implied, of the United States Air Force or the U.S. Government. The U.S. Government is authorized to reproduce and distribute reprints for Government purposes notwithstanding any copyright notation herein.

\bibliography{ref}
\bibliographystyle{icml2021}

\end{document}